# Text Summarization in the Biomedical Domain

Milad Moradi [1,2], Nasser Ghadiri [1]

[1] *Department of Electrical and Computer Engineering, Isfahan University of Technology, Isfahan 84156-83111, Iran*

[2] *Institute for Artificial Intelligence and Decision Support, Center for Medical Statistics, Informatics, and Intelligent Systems, Medical University of Vienna, Vienna, Austria*

**ABSTRACT**

*This chapter gives an overview of recent advances in the field of biomedical text summarization. Different types of challenges are introduced, and methods are discussed concerning the type of challenge that they address. Biomedical literature summarization is explored as a leading trend in the field, and some future lines of work are pointed out. Underlying methods of recent summarization systems are briefly explained and the most significant evaluation results are mentioned. The primary purpose of this chapter is to review the most significant research efforts made in the current decade toward new methods of biomedical text summarization. As the main parts of this chapter, current trends are discussed and new challenges are introduced.*

# INTRODUCTION

The large volume of textual information in the biomedical domain is always a challenge that leads researchers to develop new domain-specific text processing tools. In recent decades, automatic biomedical text summarization methods have been widely investigated to provide clinicians and researchers, generally users in the biomedical domain, with tools that help them to deal with large amounts of information embedded in textual resources.

Biomedical information is available in the form of different types of documents. Biomedical literature provides clinicians and researchers with a valuable source of knowledge to assess the latest advances in a particular field of study, develop and validate new hypotheses, conduct experiments, and interpret their results [1]. Clinical trials, medical records, multi-media documents, information on the web, and so on are other resources that contain huge amounts of valuable information [2]. The size of these textual resources is overgrowing, and it is becoming harder to extract and manage the information embedded in large available documents [3]. It is crucial in both academia and industry to develop automatic tools that facilitate exhausting tasks in the pipeline of information extraction and knowledge discovery from textual resources.

In recent decades, many automatic methods have been developed to deal with the difficulties of exploiting text documents for information extraction and knowledge discovery tasks. The methods have led to substantial advances in various crucial fields such as gene and genome expression, drug-target discovery, drug repositioning, identifying advert events, and building domain-specific databases [1]. Text mining and Natural language processing methods play an essential role in developing automatic text processing tools. Automatic text summarization is a promising approach to effective extraction and management of gainful information contained in large and lengthy text documents.

So far, many text summarization methods have been proposed to address various challenges related to different types of text documents in the biomedical domain [3, 4]. In a broad categorization, the approaches to biomedical text summarization can be divided into four classes of statistical, natural language processing, machine learning, and hybrid methods [3]. Since every type of document has its properties, it depends on the input text and its characteristics, also the task at hand, that which class of methods can be more suitable for a specific problem. The problems addressed by the summarization methods cover a wide range of subfields in the biomedical domain. Summarization of biomedical literature [2, 5-9], summarization of treatments [10], evidence-based medical care [11], summarization of drug information [12], clinical decision support [13], summarization of clinical notes [14] and electronic health records [15] are among various applications of text summarization in the biomedical domain.

This chapter gives a review of recent advances in the field of biomedical summarization. Since researchers in this field always identify new challenges and address them by adopting novel approaches, it can be essential to review the state-of-the-art. This can help to get familiar with new challenges and problems, the most efficient approaches, and the most significant results obtained from evaluation methods. In overall, this survey can provide an overview of the recent research that pushes the boundaries of biomedical text summarization.

The comprehensive review presented in this chapter may introduce some new challenges that have not been addressed so far. Hence, the chapter can be a good start point for those who intend to start researching the field of biomedical text summarization.

## BACKGROUND

Text summarization methods can be categorized into two classes of abstractive and extractive [16]. In abstractive summarization, natural language understanding and generation techniques are used to interpret the input text and generate a new version that is shorter and conveys the main ideas. On the other hand, an extractive summarizer does not need to produce a new text. It identifies the most important ideas within the text and extracts those parts of the text that are highly relevant to the main ideas. The majority of studies in the field of biomedical text summarization focus on extractive methods since dealing with difficulties of abstractive summarization needs comprehensive knowledge in various subfields of linguistics and natural language processing.

Concerning the number of inputs, summarization methods are divided into two categories, i.e., single-document and multi-document [3]. Dealing with multiple inputs is more challenging because important information is distributed among a set of potentially heterogeneous documents. Reduction of redundant information is also a severe challenge that needs to be addressed. Since sentences are put together from different documents, cohesion and reference resolution can be other problems in multi-document summarization.

Generic versus user-oriented summarization is another classification [3]. In generic summarization, the system identifies important topics within the input text and produces a summary that covers those critical ideas. On the other hand, in user-oriented summarization, the user gives the system his preferences in the form of a query or a set of keywords; and the system generates a summary that addresses requirements of the user. There are also additional classifications for summarization systems, such as supervised, unsupervised, informative, indicative, and so on. Figure 1 presents the general criteria in categorizing text summarization systems.

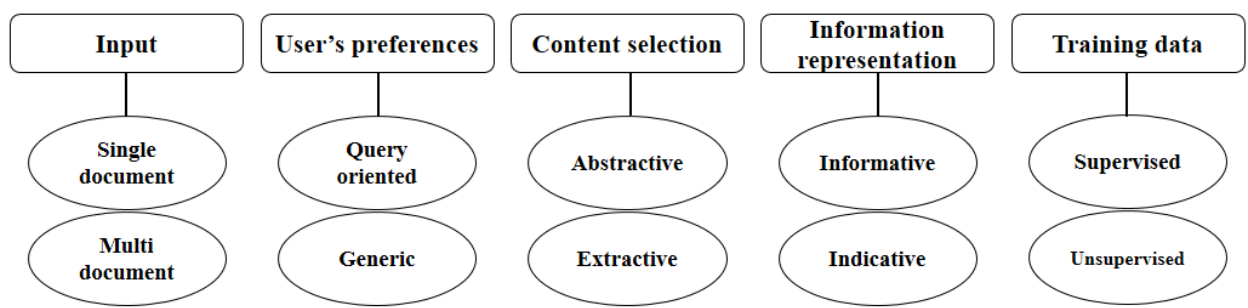

*Figure 1. Different criteria in categorizing text summarization methods*

Due to some limitations of general-purpose methods for summarization of biomedical texts, there has been a tendency to developing domain-specific summarizers through utilizing sources of domain knowledge in the summarization process. This has led to significant improvements in the performance of biomedical summarization [3]. Sources of biomedical knowledge are available in different forms of controlled vocabularies, taxonomies, ontologies, or some combinations of them. A controlled vocabulary contains organized sets of words and phrases related to a particular field. Controlled vocabularies are considered as means of organizing knowledge, and are widely used for content indexing and retrieval. A taxonomy is used to classify concepts into groups based on their similarities, differences, or other criteria. For example, a taxonomy might serve as a classification system by grouping diseases based on involved body organs or any other classification criteria. An ontology contains a set of concepts, definition and categorization of those concepts, and relationships between them. It can be used as a knowledge representation tool in information processing systems that need to interact with sources of domain knowledge of a particular topic.

The Unified Medical Language System (UMLS) [17] has been widely used in a variety of biomedical natural language processing tasks, especially in summarization. It integrates more than 100 biomedical vocabularies and ontologies into three main components, i.e., Metathesaurus, Specialist Lexicon, and Semantic Networks. These components contain large amounts of lexicographic information, biomedical concepts, and their semantic relationships. Medical Subject Heading (MeSH) is a controlled vocabulary covering a wide range of terms and phrases in the life sciences. It organizes terms into a hierarchical terminology that can be utilized in systems that rely on indexing, searching, and retrieval of biomedical information. MeSH also records definitions, descriptors, and synonyms of terms.

This chapter covers a wide range of summarization methods developed to address different tasks such as biomedical literature summarization, automatic abstract generation, developing decision support tools, biomedical data curation, and so on. Methods fall into different categories of abstractive, extractive, single-document, multi-document, generic, and user-oriented. Most of the systems incorporate sources of domain knowledge in different stages of the summarization process, and the most widely used knowledge source is the UMLS.

## RECENT ADVANCES IN BIOMEDICAL TEXT SUMMARIZATION

### Summarization of biomedical literature

A graph-based approach to biomedical summarization [6] is one of the comprehensive works toward summarizing scientific biomedical articles. It converts the input text to a graph representation in which the concepts extracted from the UMLS constitute the nodes, and the semantic relationships between the concepts form the edges. A clustering algorithm is employed to divide the nodes and edges into a set of subthemes denoting main topics of the text. Different heuristics are evaluated for sentence selection. When each cluster contributes to the summary in proportion to its size, the system reports the highest summarization scores. The impact of generic features, such as position and similarity to the title, are also evaluated for the sentence selection stage.

An investigation of the impact of different sources of domain knowledge is done with the use of semantic graph-based text modeling [18]. It shows that the performance of literature summarization can improve when appropriate knowledge sources are utilized to represent the input text by a concept-based model. Specificity and extensive coverage of concepts and semantic relations are mentioned as characteristics of an appropriate knowledge source.

A previous study [19] utilizes MeSH terms as the basis of text reduction for document retrieval and information extraction. The goal is to mediate between the extremes of abstracts and full-texts. The method assumes the MeSH terms used for indexing a document are reliable indicators of the most important ideas within the document. Some similarity functions assess the relatedness of sentences to the MeSH terms, and the most related ones are extracted to form the summary. The evaluations show some degrees of correlation between system-produced summaries and human judgments.

The Itemset-based summarizer [2] is the first method in biomedical text summarization that uses frequent itemset mining for extracting main topics and ideas within an input text. It introduces the quantification of informative content for the semantics behind the words and sentences. The summarization process begins by a preprocessing stage in which the input text is mapped to the UMLS concepts. The text is represented in a transactional format, then in the topic extraction stage frequent itemsets are discovered. The sentences are scored based on the presence of the main topics, and the most informative ones are selected to build the summary.

After proposing the Itemset-based summarizer, other research works have exploited the use of frequent itemsets for extracting main topics of biomedical texts. A graph-based approach [20] is among the methods that utilize frequent itemsets to map the input text to an intermediate representation. In this method, a similarity measure approximates the similarity between each pair of sentences based on the frequent itemsets that the sentences have in common. Then a graph is constructed by considering the sentences as the nodes and the similarity values as the weights. A minimum spanning tree clustering algorithm divides the sentences into a set of clusters. At the final stage, the most important sentences of every cluster are selected to form the summary.

Clustering and Itemset mining based Biomedical Summarizer (CIBS) is the latest effort toward exploiting frequent itemsets in biomedical summarization [7]. CIBS addresses the challenges related to information coverage and redundancy in multi-document summarization. After extracting frequent itemsets as the main subtopics, CIBS employs a hierarchical clustering algorithm to divide the sentences into multiple clusters. CIBS also introduces a measure to approximate the extent of important information covered by each pair of sentences or clusters. Those sentences within the same cluster cover a set of main topics that are not covered (or partially covered) by the sentences within other clusters. The challenge of establishing a trade-off between information coverage and redundancy is discussed in both single- and multi-document summarization.

Recent work in biomedical literature summarization addresses the challenges related to statistical and probabilistic methods [5]. It proposes a heuristic method based on the Naïve Bayes classification paradigm to classify the sentences into two classes of summary and non-summary. The underlying assumption followed to estimate the probabilities is that the distribution of essential concepts within the summary should be similar to the distribution within the original text. It is also discussed that there are other measures, rather than the simple frequency, that can be used in probabilistic summarization. Various approaches are evaluated for feature selection, and some of them achieve better performance than the simple frequency method. The meaningfulness and Concept Frequency-Invert Paragraph Frequency (CF-IPF) measures report significant improvement. It is also shown that when the correlations between concepts are embedded in the feature selection step, the summarizer can more efficiently identify the most relevant content.

Different similarity measures are assessed for their usefulness in a graph-based approach to summarization of biomedical literature [21]. Four similarity measures, i.e. Cosine, Jaccard, TextRank, and positional similarity, are used to approximate the similarity between sentences in terms of concepts and semantic types extracted from the UMLS. The similarity values are considered as weights in the graph constructed for the input text. A clustering algorithm divides the sentences into groups, and the summary is produced by selecting multiple sentences from every cluster. It is shown that when both concepts and semantic types are used for assessing the similarity between sentences, the summarizer obtains higher scores for all the similarity measures.

The sentence position has been a widely-used feature in text summarization and other natural language processing tasks. This feature performs well for summarizing specific types of documents such as news articles. However, a study on the usefulness of positional features [22] demonstrates that the traditional sentence position feature should be reinforced to achieve desirable performance in biomedical literature summarization. It is shown that when different weights are assigned to sentences based on their position within the article, the summarizer can produce better summaries than the strategy that assigns weights to only sentences appearing at beginning or end of the article.

Semantic relation extraction is also investigated for summarization of scientific articles toward specific biomedical concepts [23]. This system consists of three stages. Using the SemRep tool, semantic relations are extracted in the first stage. Next, in the second stage, several sets of sentences are identified and divided into separate groups based on the semantic relations appearing in each sentence. Finally, the most informative sentences of each set are retrieved to cover all types of semantic relations in the final summary. The system reports a significant improvement in the performance of multi-document biomedical summarization in comparison to the traditional MEAD method.

Some biomedical literature summarization methods address more fine-grained problems. Among them, the task of identifying citation sentences is addressed by a summarization technique based on Support Vector Machine (SVM) classification [24]. This type of summarization aims at extracting sentences within an article that refers to other articles. This helps the user to investigate complimentary or contradictory materials for a given article. This system combines different tasks such as feature extraction, sentence classification, and rule-based post-processing to generate the final output that is a summary in the form of citation sentences.

**Summarization of MEDLINE abstracts**

MEDLINE records more than 20 million scientific articles from life sciences and the biomedical domain. It can be used as a valuable source of information in a wide range of biomedical topics. Searching in the massive volume of abstracts stored in MEDLINE can result in retrieving a large number of documents. Hence, some efforts have been made toward summarizing this type of biomedical texts. A graph-based approach to summarization of treatment of diseases uses abstracts retrieved by PubMed as the input text [10]. It creates a graph for four types of clinical concepts related to four aspects of treatments, i.e. location, drugs, comorbidities, and procedures. The resulted graph is exploited to extract the most crucial aspects of treatment and produce the final summary.

Semantic MEDLINE [25] is another tool that provides decision support data through summarization of MEDLINE abstracts. It acts based on semantic predications of SemRep extracted for some particular concepts specified by the user. The system is proposed for two types of conventional and dynamic summarization. In a general summary, five classes of information are identified, i.e., diagnosis, genetic etiology of disease, pharmacogenomics, substance interaction, and treatment of disease. Semantic predications concerning these classes of information are refined through four filters of relevance, connectivity, novelty, and saliency. The dynamic method utilizes a dynamic statistical algorithm to perform the saliency measurement in an online manner. This leads to having more accurate predications with respect to the other tree filters. Finally, the predications help the system to produce decision support data.

Other work on summarization of MEDLINE abstracts [26] adopts a graph-based approach relying on clique clustering. It extracts SemRep predications and constructs a predication graph. It then applies three filters of novelty, centrality, and frequency to identify cliques. Finally, the summary is produced, and cliques are clustered to reveal the main themes in the summary. Since valid clusters can improve the quality of summaries, the study also evaluates the utility of clusters in terms of measures of cohesion, separation, and overall validity.

**Automatic abstract generation**

COMPENDIUM [27] is an automatic abstract generation system aiming at both extractive and abstractive summarization of biomedical texts. The extractive method utilizes a set of natural language processing stages such as surface linguistic analysis, redundancy detection, topic identification, and relevance detection. In the redundancy detection stage, it uses a textual entailment approach helping to omit repeated contents. The traditional term frequency feature is employed for identifying important topics. For the relevance detection task, COMPENDIUM incorporates the Code Quantity Principle and makes use of frequent words to discover the most informative sentences.

For an abstractive summary generation, COMPENDIUM integrates the extractive method with a stage in which information compression and fusion tasks are performed. These tasks are accomplished through a set of stages, i.e., word graph generation, incorrect path filtering, and combining given and new information.

The system is evaluated for both the extractive and abstractive summarization in terms of quantitative and qualitative criteria. Both the approaches are able to include relevant information. However, the abstractive method performs better concerning user satisfaction assessment.

**Facilitating evidence-based practice**

Comprehensively screening high- quality studies can facilitate evidence-based practice. The screening process can result in producing systematic reviews that are valuable sources of evidence. However, this process is highly time- and labor-intensive. An automatic tool [28] is proposed that represents the similarities between articles and summarizes their content into a semantic space that can facilitate the article screening and literature review tasks. The semantic space is constructed by mapping the input text to concepts and semantic relations extracted from the UMLS. This system improves the performance on the task of identifying relevant articles to a collection of systematic reviews, compared to lexical features and corpus-based semantic approaches. This type of summarization can be considered indicative since the summary refers to essential contents of the text, in contrast to the informative approach that the summary contains the critical parts and the user does not need to refer to the original text.

Other summarization systems [29] specialized for evidence-based medicine receives research abstracts of randomly-controlled trials as inputs and produces a summary statistics. The system searches in descriptions of the treatment groups and outcomes, also other properties related to a clinical trial, and calculates summary statistics related to two standard measures of effectiveness of interventions, i.e., absolute risk reduction and number needed to treat. This type of summary statistics can significantly decrease the resources needed to seek for the latest research findings.

**Summarization as a tool for data curation**

Text summarization methods can be highly specialized to address specific information retrieval needs. An effort toward developing such tools is Semantic MEDLINE that facilitates the task of automatic information extraction from the biomedical literature. A system [30] is developed as a component of Semantic MEDLINE and specialized for summarizing information related to molecular genetics within a collection of text documents. The system relies on the predications extracted from the SemRep; and is proposed as an assistant for data curators, helping them with building secondary databases of genetic information.

**Resolving ambiguity in mapping text to concepts**

Some methods have improved the performance of biomedical summarization by modeling the input text as a network of connected concepts extracted from the UMLS [6, 31]. However, in many cases, there may be multiple concepts returned for a word or phrase due to ambiguous terms. In this situation, it may be needed to resolve the ambiguity to avoid degradation in the accuracy of the model. An initial work [32] demonstrates that when an effective disambiguation strategy addresses lexical ambiguity, the quality of summaries produced for biomedical text documents is enhanced.

Another study [33] investigates the accuracy of word sense disambiguation methods intrinsically on the NLM WSD dataset and the MSH WSD dataset, also extrinsically as a part of a biomedical text summarizer. The results demonstrate that those disambiguation methods performing better in intrinsic evaluation can also obtain better scores in extrinsic experiments as a stage in the summarization pipeline.

Different strategies were evaluated to resolve disambiguation in a graph-based biomedical summarizer [31]. It was shown that when all the candidate concepts are considered for building a graph from the text, the model can comparably produce acceptable results. Furthermore, when all the candidate mappings are weighted using some weighting algorithms, the summarization system can achieve its best performance. Other strategies such as Personalized PageRank and JDI report lower scores for resolving ambiguity in this type of summarization.

**FUTURE RESEARCH DIRECTIONS**

As discussed in this chapter, there have been significant advances in developing biomedical text summarization methods in recent years. There are different types of challenges for various types of documents available in the biomedical domain. However, the majority of methods are evaluated for summarization of biomedical literature. This is because a vast number of scientific articles are freely accessible in either form of abstract or full-text.

Furthermore, there are some limitations with accessing to other types of documents like medical records or clinical trials. The utility of novel summarization methods can be comprehensively investigated by conducting experiments on as various types of documents as possible. This needs much effort to collect and process large numbers of text documents and their model summaries. By focusing on more task-specific summarization methods, future studies can go beyond literature summarization and lead to developing datasets and benchmarks for other types of documents not addressed so far.

A significant limitation in the evaluation of biomedical summarizers may be the lack of any standard datasets or benchmarks. For literature summarization, the existing studies randomly select some articles from publicly available corpora, especially PubMed, and use abstracts of the articles as model summaries. This always leads to controversies in the community because the abstract is a summary from the author's perspective and may not convey important parts of the full-text indeed. Also, the abstract does not contain exact words and sentences that appear in the full-text. This can negatively affect the reliability of evaluation results because the evaluation metrics act based on the content overlap between system-produced and model summaries with respect to the same words appearing in both summaries. Therefore, an extensive effort is needed toward developing a standard dataset and benchmark addressing the above-mentioned challenges.

The evaluation of summarization methods can be done through two approaches of intrinsic and extrinsic. In the intrinsic evaluation, the performance of the summarizer is directly assessed regarding the quality of summaries generated by the system. On the other hand, extrinsic approach evaluates the utility of summarization by the improvement obtained in terms of other tasks such as information retrieval and extraction, or decision making. The majority of evaluation in the field has been done through the intrinsic approach. It will be valuable to go beyond assessing the usefulness of summarizers just based on the quality of summaries. The reason may be that the user often needs the summary as a peripheral mean to accomplish other tasks like information retrieval and extraction. Subsequently, by embedding summarization methods as a peripheral task into the pipeline of other primary tasks, and conducting the extrinsic evaluation, summarizers can be developed to address real-world challenges.

As discussed earlier in this chapter, many summarization methods exploit sources of domain knowledge to map the input text to a semantic representation. This can help methods to approximate semantics behind words and sentences, leading to more accurate representations of documents. This has led to the current trend of developing knowledge-rich methods. The UMLS is the most common source of domain knowledge utilized by a variety of summarization systems. However, the choice of proper knowledge source is still an open challenge that needs more attention from the research community. Models of summarization can be more accurate by utilizing task-specific sources of domain knowledge or combining existing resources. It has been shown that when the importance of sentences is assessed based on their content rather than the generic features, the quality of summaries can improve. The next generation of knowledge-rich methods may benefit from neural network-based language models since they have shown their superiority for capturing the context in which different parts of text appear. Furthermore, neural network-based language models provide generalizability over a wide range of natural language processing tasks. This can improve the performance of systems in which a pipeline of multiple natural language processing tasks, including summarization, work together; and a single generalizable model can improve the accuracy and decrease the cost.

Many previous works address the summarization of Electronic Health Record (EHR) data as a valuable source of biomedical information. However, most of the attention is paid to visualization of data; and text summarization acts as a peripheral task. Future research may include combining literature and EHR summarization to help with clinical decision support. A huge volume of knowledge in the biomedical domain is accessible through the scientific literature, and as shown by previous work, this knowledge can be exploited in clinical settings with the use of summarization systems. Adopting novel techniques from data analysis and data fusion, hybrid systems may be devised to summarize the scientific literature in combination with medical records with the goal of developing a new hypothesis, inferring new knowledge, and building new domain-specific databases.

**CONCLUSION**

This chapter presented a review of recent advances in biomedical text summarization. It was discussed that there had been a trend toward devising systems that incorporate domain knowledge to enhance the accuracy of text modeling. An overview of the most common tasks that utilizes text summarization was presented. It was shown that most of the studies address the challenges related to the summarization of biomedical literature. However, it is still needed to create standard benchmarks to allow interpreting results concerning regular evaluations.

The majority of methods focus on summarization of biomedical literature. It seems that there is much potential to develop more task-specific methods to exploit text summarization as a stage in large pipelines of information retrieval, knowledge discovery, and decision making systems.

**KEY TERMS AND DEFINITIONS**

**Clustering:** A machine learning method that groups data records into a set of clusters, such that each record has the maximum similarity to records within the same cluster and the minimum similarity to the records within other clusters.

**Decision Support System:** A computer system that facilitates the process of decision making by gathering, storing, analyzing, and visualizing information.

**Generic Feature:** A feature in text summarization methods that refers to general properties of sentences like length and position, regardless of the semantics behind the words.

**Indicative summarization:** An indicative summary only contains some indicators to the important parts of the input, and the user needs to refer to the original text to read more explanations.

**Informative summarization:** An informative summary directly conveys important parts of the input text instead of only containing indicators to those parts.

**Itemset Mining:** A data mining technique that discovers correlated data items within a large dataset. It also produces measures of support and confidence to show the strength of evidence in favor of a correlation.

**Positional Feature:** A feature in text summarization methods that refer to the relative location of a sentence in a text document.

**Supervised Summarization Method:** A class of summarization methods in which the summarizer is provided with a set of training data to learn patterns of relationship between features and some labels assigned to the sentences. The labels specify that which sentences should be selected or should not be selected for inclusion in the summary.

**Unsupervised Summarization Method:** A class of summarization methods in which there is no training data; and the summarizer should decide which sentences are the most important and relevant based on a set of features or the content of the input text.